\documentclass[a4paper]{jpconf}
\usepackage{graphicx}
\usepackage{amsmath}
\usepackage{hyperref}
\usepackage{cleveref}
\usepackage{comment}
\usepackage{multirow}
\usepackage{amssymb}
\usepackage{booktabs}
\usepackage{gensymb}
\usepackage{makecell}
\usepackage{array}

\usepackage{tabularx}
\usepackage{arydshln}
\newcommand{\specialcell}[2][c]{%
  \begin{tabular}[#1]{@{}c@{}}#2\end{tabular}}
\usepackage{pifont}
\usepackage{xcolor}

\definecolor{darkgreen}{RGB}{0, 75, 0}
\definecolor{darkgray}{RGB}{169, 169, 169}
\definecolor{orange}{RGB}{200, 120, 0}
\definecolor{teal}{RGB}{0, 128, 128}

\begin{document}
\title{Thermoxels: a voxel-based method to generate simulation-ready 3D thermal models }

\author{Etienne Chassaing,$^1$ Florent Forest,$^1$ Olga Fink,$^1$ Malcolm Mielle$^2$}

\address{$^1$ IMOS, École Polytechnique Fédérale de Lausanne (EPFL), Lausanne, Switzerland

  $^2$ Schindler EPFL Lab, Lausanne, Switzerland}

\ead{malcolm.mielle@ik.me}

\begin{abstract}
  In the European Union, buildings account for 42\% of energy use and 35\% of greenhouse gas emissions.
  Since most existing buildings will still be in use by 2050, retrofitting is crucial for emissions reduction.
  However, current building assessment methods rely mainly on qualitative thermal imaging, which limits data-driven decisions for energy savings.
  On the other hand, quantitative assessments using finite element analysis (FEA) offer precise insights but require manual CAD design, which is tedious and error-prone.
  Recent advances in 3D reconstruction, such as Neural Radiance Fields (NeRF) and Gaussian Splatting, enable precise 3D modeling from sparse images but lack clearly defined volumes and the interfaces between them needed for FEA.
  We propose Thermoxels, a novel voxel-based method able to generate FEA-compatible models, including both geometry and temperature, from a sparse set of RGB and thermal images.
  Using pairs of RGB and thermal images as input, Thermoxels represents a scene's geometry as a set of voxels comprising color and temperature information.
  After optimization, a simple process is used to transform Thermoxels' models into tetrahedral meshes compatible with FEA.
  We demonstrate Thermoxels' capability to generate RGB+Thermal meshes of 3D scenes, surpassing other state-of-the-art methods.
  To showcase the practical applications of Thermoxels' models, we conduct a simple heat conduction simulation using FEA, achieving convergence from an initial state defined by Thermoxels' thermal reconstruction.
  Additionally, we compare Thermoxels' image synthesis abilities with current state-of-the-art methods, showing competitive results, and discuss the limitations of existing metrics in assessing mesh quality.

\end{abstract}

\section{Introduction}

Buildings account for a large part of energy use and carbon emissions---in 2020, buildings represented 40\% of the total energy consumption in the European Union (EU) and 36\% of EU greenhouse gas emissions~\cite{european2020renovation}.
In the EU, it is forecasted that 85-95\% of existing buildings will still be standing in 2050, and while most of those buildings are not energy-efficient,
only 11\% of the existing building stock undergoes some level of renovation each year.
Furthermore, most renovation efforts do not address the energy performance of buildings.

While the need for deep energy-efficient renovation is clear, evaluating the impact of renovations is complex and time-consuming.
Quantitative evaluation of a given building's thermal envelope requires experts to conduct on-site measurements and manual design of volumetric models compatible with finite element analysis (FEA) simulation software.
Such volumetric models are often generated from computer-assisted design models, a process that is both error-prone and time-consuming~\cite{sanhudo_building_2018}.
%
On the other hand, recent advances in photogrammetry and 3D computer vision methods---such as Structure-from-Motion (SfM), Neural Radiance Fields (NeRF)~\cite{mildenhall2020nerf}, and 3D Gaussian Splatting (3DGS)~\cite{chen_survey_2024}---enable precise 3D modeling from photographs.
However, the resulting models lack clearly defined volumetric elements (point clouds and NeRF)~\cite{wang_neus_2021,xu_exploiting_2025}
and interfaces (3DGS), needed for FEA~\cite{xie_physgaussian_2023}.


In this paper, we introduce Thermoxels, a novel end-to-end method for generating 3D RGB+thermal models that are FEA-compatible, using only a sparse set of RGB and thermal photographs.
Thermoxels jointly optimizes geometry and temperature properties at the voxel level to produce volumetric meshes suitable for simulation.
Our main contribution is a framework to generate thermal voxel-based models that are simulation-compatible and with accurate temperature distributions.

\section{Related work}

\textbf{Finite Element Analysis (FEA) and Building Information Modeling (BIM) for building retrofit.}
While retrofitting of existing building infrastructure is a key factor in reducing carbon footprint, it is also a complex task that often relies on expert knowledge and qualitative assessments~\cite{sanhudo_building_2018}.
On the other hand, precise quantification of the energy savings produced by a retrofit requires running FEA on an accurate, FEA-compatible building model.
FEA is the process of predicting an object's physical behavior based on calculations made with the finite-element method (FEM) from a 3D model; for thermal analysis, it necessitate both geometric information and thermal properties~\cite{carrasco_building_2023}.
This model is typically created via manual CAD modeling or using Building Information Modeling (BIM) software, which enables the creation of a discretized 3D model of the building, including geometry, thermal properties, and boundary conditions.
However, creating a BIM model is a complex and time-consuming process that requires expert knowledge or expensive sensors~\cite{cho_3d_2015}.

\noindent\textbf{3D reconstruction and the built environment}.
Neural Radiance Fields (NeRF)~\cite{mildenhall2021nerf} use a set of multi-layer perceptrons (MLPs) to implicitly represent the 3D geometry of the scene and enable the rendering of novel views from any point in space. To improve surface reconstruction, a signed distance function (SDF) can be used~\cite{xu_exploiting_2025}.
On the other hand, 3D Gaussian Splatting (3DGS)~\cite{kerbl20233d} forgoes the use of MLPs, and instead uses a set of 3D Gaussians to represent the geometry and color of the scene.
However, none of the methods can generate a mesh compatible with FEA without extensive manual post-processing.
In contrast, voxel-based representations~\cite{fridovich2022plenoxels, li2022vox} divide the space into a grid of voxels, and the value of a given pixel is interpolated from the values of the voxels that intersect the ray cast from it during rendering.
While NeRF-based~\cite{hassan_thermonerf_2024} and 3DGS-based~\cite{chen2024thermal3d} multimodal RGB+thermal models have been proposed, voxel representations are better suited for FEA.
However, the potential of these models to create RGB+thermal representations compatible with FEA, particularly in the context of thermal simulations, has not yet been explored.


\noindent\textbf{Thermal computer vision}.
Conducting FEA on a 3D building model requires obtaining the initial temperature distribution as an initial condition.
Temperature can be measured using thermal cameras, which measure thermal radiation.
However, unlike near-infrared sensors operating near the visible light spectrum, thermal cameras detect thermal radiation scattered and reflected by objects and their surroundings.
This process often results in thermal images that appear soft and lack texture, a phenomenon referred to as the ghosting effect.
Due to the ghosting effect, early studies on 3D scene reconstruction from thermal images have used RGB images to reconstruct the scene and projected the thermal information on the 3D model afterward, possibly leading to discrepancies between the geometry and thermal information due to incorrect alignment~\cite{zeise_temperature_2016,hassan_thermonerf_2024}.


\section{Method}

\begin{figure}
  \includegraphics[width=\textwidth]{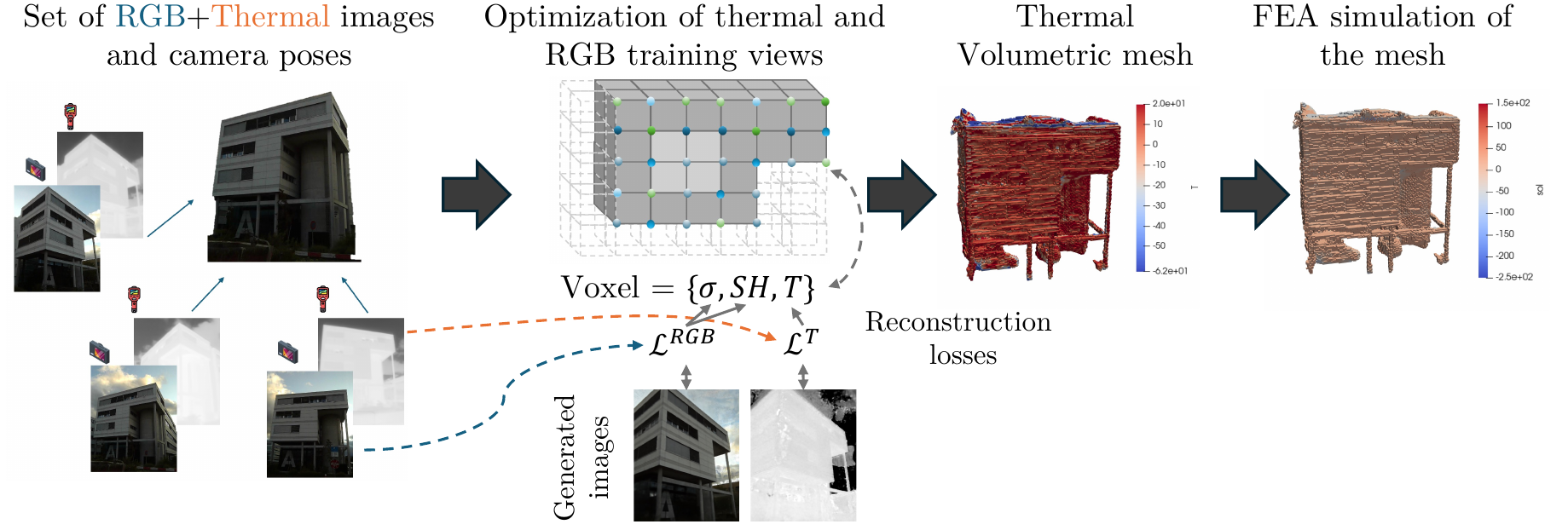}
  \caption{Pipeline of the proposed Thermoxels framework transforming input images into a simulation-ready 3D thermal model.
    Given a set of RGB and thermal images, the density ($\sigma$), color (spherical harmonics $SH$), and temperature ($T$) of a voxel-based 3D model are optimized by comparing synthesized RGB and thermal images (based only on camera pose) against ground-truth measurements.
    Images below the voxel grid are Thermoxels-generated views on the Building A spring dataset.
  }
  \label{fig:summary}
\end{figure}


We propose a novel voxel-based architecture that includes density, RGB, and thermal information, in one consistent 3D model.
To achieve this, we use paired RGB and thermal images: the RGB images supervise the learning of color and density fields, while the thermal images are used exclusively to optimize the temperature field.
Once the voxel-based model is optimized, the estimated density is used to create a volumetric mesh compatible with FEA.
For a visual representation of our method, refer to \cref{fig:summary}.


\subsection{Thermoxels}

Thermoxels represents the space as a voxel grid, where, for all voxels, the corner contains a density value $\sigma$, spherical harmonics $SH$ coefficients, and a temperature value $T$:
$
  V_i = \{ \sigma_i, SH_i, T_i \}
$
To obtain the color and temperature of a given pixel, a ray $r$ passing by the pixel is traced in space based on the camera parameters, and $N$ positions $p_i = (x, y, z)$ are sampled at regular intervals along $r$.
Similarly to Plenoxels~\cite{fridovich2022plenoxels}, the color $c_i$ of each $p_i$ is determined by trilinear interpolation of the nearest spherical harmonics, enabling the rendered image to exhibit continuous color variations despite the discrete nature of the representation.
The final color $\hat{C}$ for the pixel is computed by integrating the contributions of each sample along $r$:
\begin{equation}
  \hat{C}(r) = \sum_{i=1}^{N}\alpha_i (1 - \exp(-\sigma_i \delta_i)) c_i,
  \quad \text{with} \, \,
  \alpha_i = \exp ( -\sum_{j=1}^{i-1} \sigma_j \delta_j ).
\end{equation}
where $(1 - \exp(-\sigma_i \delta_i))$ is the light contribution from sample $i$, $\sigma_i$ the density of sample $i$, and $\delta_i$ the distance between samples.
$\alpha_i$ is the accumulated transmittance along the ray up to the $i^{th}$ sample, and $\delta_i$ is the distance between the $i^{th}$ and $(i+1)^{th}$ samples.

Similarly, the temperature value $T_i$ of a given $p_i$ is obtained by trilinear interpolation between the temperature of the 8 closest corners.
However, unlike colors, temperatures are not represented by spherical harmonics; instead, temperatures are fixed values at each corner, representing a unique surface temperature, independent of viewpoint.
Hence, when using interpolation, we assume gradual temperature changes in space---given a small enough resolution of the grid, this assumption is reasonable thanks to temperature field continuity.
The final temperature $\hat{T}$ of a pixel is estimated by integrating the temperatures of all $p_i$ on a ray $r$:
\begin{equation}
  \hat{T}(r) = \sum_{i=1}^{N}\alpha_i (1 - \exp(-\sigma_i \delta_i))T_i
\end{equation}

It should be noted that, while for RGB rendering the background colors---i.e., pixel rays that only go through empty voxels---are estimated through a spherical background image, in the thermal rendering no background is learned and background pixels are instead set to the minimum temperature of the scene for rendering.

To optimize the model, synthetic images are rendered from the voxel grid using the above procedure and compared to ground-truth images.
In our experiments, the temperatures of the ground-truth thermal images are normalized between 0 and 1,
and the initial temperature of each voxel corner is set to 0.5$\degree$C before optimization.
The final loss function of Thermoxels, including both RGB and temperature losses, is:
\begin{equation}
  \mathcal{L} = \mathcal{L}_2^{rgb} + \lambda_{TV}^{rgb} \mathcal{L}_{TV}^{rgb} + \lambda \mathcal{L}_2^T + (1-\lambda) \mathcal{L}_1^T + \lambda_{TV}^{T} \mathcal{L}_{TV}^T
\end{equation}
with $\mathcal{L}_2^{rgb}$ the L2 loss between the rendered and ground-truth RGB images and $\mathcal{L}_{TV}^{rgb}$ the total variation (TV) loss on the RGB image~\cite{fridovich2022plenoxels}.
$\mathcal{L}_2^T$ and $\mathcal{L}_1^T$ are the L2 and L1 losses between the rendered and ground truth thermal images, respectively.
Through experiments, we found that $\lambda=\frac{1}{2}$ leads to the best results on our selected metrics, and we use this value in the rest of this paper.
Finally, $\mathcal{L}_{TV}^T$ is the thermal TV loss:
\begin{equation}
  \mathcal{L}_{TV}^T = \frac{1}{|\mathcal{V}|} \sum_{v \in \mathcal{V}} \sum_{d \in [D]} \sqrt{\Delta T_x^2(v, d) + \Delta T_y^2(v, d) + \Delta T_z^2(v, d)}.
\end{equation}
with $\Delta T_x$, $\Delta T_y$, and $\Delta T_z$ the difference in temperature between neighboring voxels along the x, y, and z axis respectively.
The temperature values of each voxel are updated using backpropagation.


\subsection{Volumetric mesh extraction}
\label{sec:method:volumetric_mesh}

Once the Thermoxels voxel-based model has been optimized, the last step of our method consists of extracting a volumetric mesh from its voxel grid.
First, the density and temperature of each voxel are calculated by trilinear interpolation at the center of the voxel.
Given the set $V$ of all non-empty voxels in the model, the voxels are filtered by keeping the set of the $t\%$ densest voxels---with $t$ a scene-dependent threshold.
This filtering step can introduce disconnected voxels in the volumetric mesh; hence, the final volumetric mesh consists of the largest set of connected voxels.
The final volumetric mesh is a single volumetric structure with each voxel having a well-defined volume, interface with other voxels, and temperature, making it well-suited for FEA.
The structured grid of regular cubic elements improves the mesh quality for FEM, increasing numerical stability~\cite{chai_bladder_voxel_fem_2011}.

\section{Evaluation}

In this paper, we aim to generate voxel-based meshes compatible with FEA, and do not focus solely on the synthesized images---indeed, as will be shown in this section, being able to synthesize high-quality images does not guarantee high-quality volumetric meshes.
Hence, we evaluate Thermoxels' ability to both generate volumetric meshes for FEA (\cref{sec:eval:fea}) and synthesize novel views (\cref{sec:eval:synthesized}).
Experiments are conducted on the non-flat scenes of the ThermoScenes dataset~\cite{hassan_thermonerf_2024}, which includes both building facades and everyday scenes.
Implementation of Thermoxels and parameter values used for each scene are available online\footnote{\url{https://github.com/Schindler-EPFL-Lab/thermoxels}}.

%

\subsection{Volumetric mesh and FEA simulation}
\label{sec:eval:fea}

Due to the lack of ground-truth meshes in the ThermoScenes dataset, we qualitatively evaluate the mesh quality of Thermoxels reconstructions.
To highlight the benefits of using multi-modal inputs, we compare Thermoxels meshes with those built by Plenoxels$_{\text{t}}$, a thermal-only adaptation of Plenoxels~\cite{fridovich2022plenoxels} without RGB supervision.

\begin{figure}[t]
    \centering
    \resizebox{\textwidth}{!}{\begin{tabular}{cccccc}
    \toprule
    Scene & Cup & Kettle & \makecell{Building \\A spring} & Dorm 1 & Dorm 2 \\
    \midrule
        \makecell{Plenoxels$_t$} &
        \makecell{\includegraphics[height=0.1\textwidth]{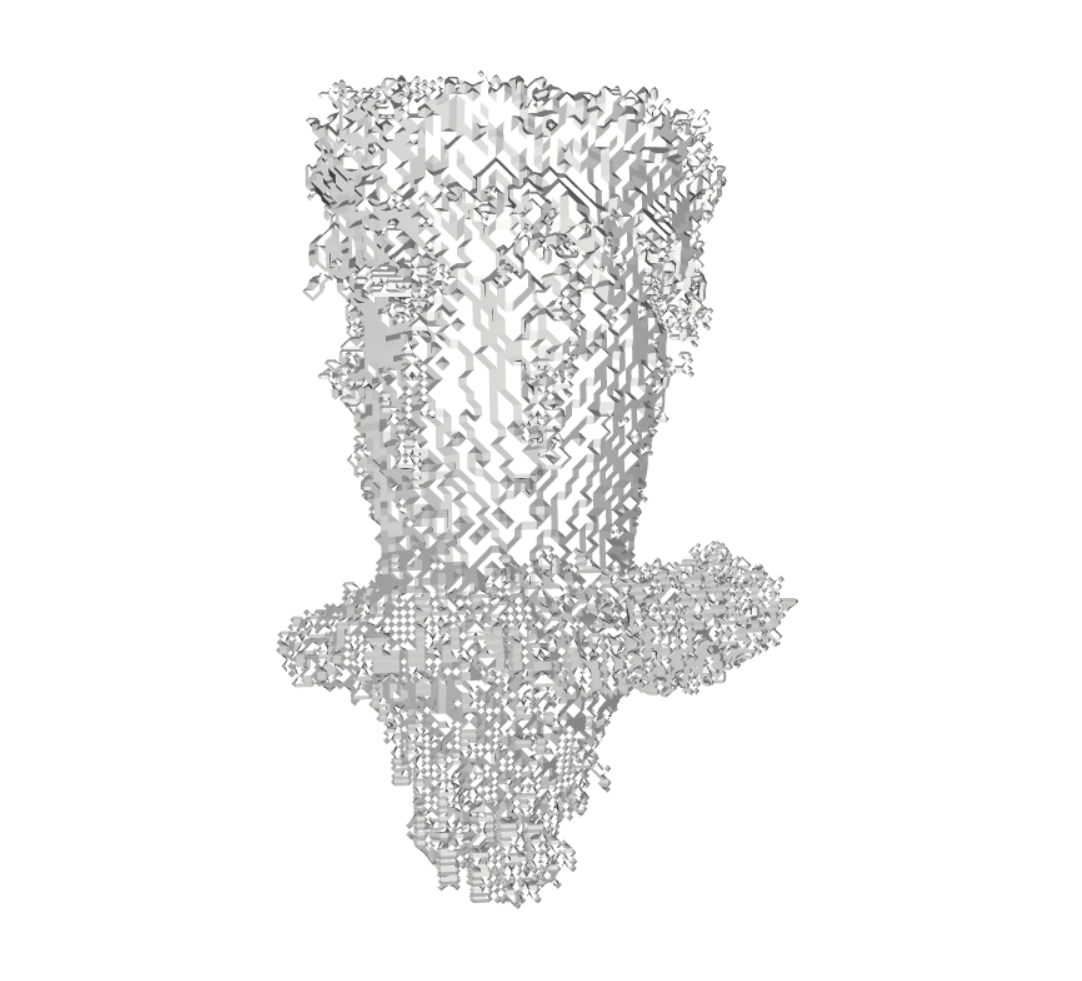}}       &
        \makecell{\includegraphics[height=0.1\textwidth]{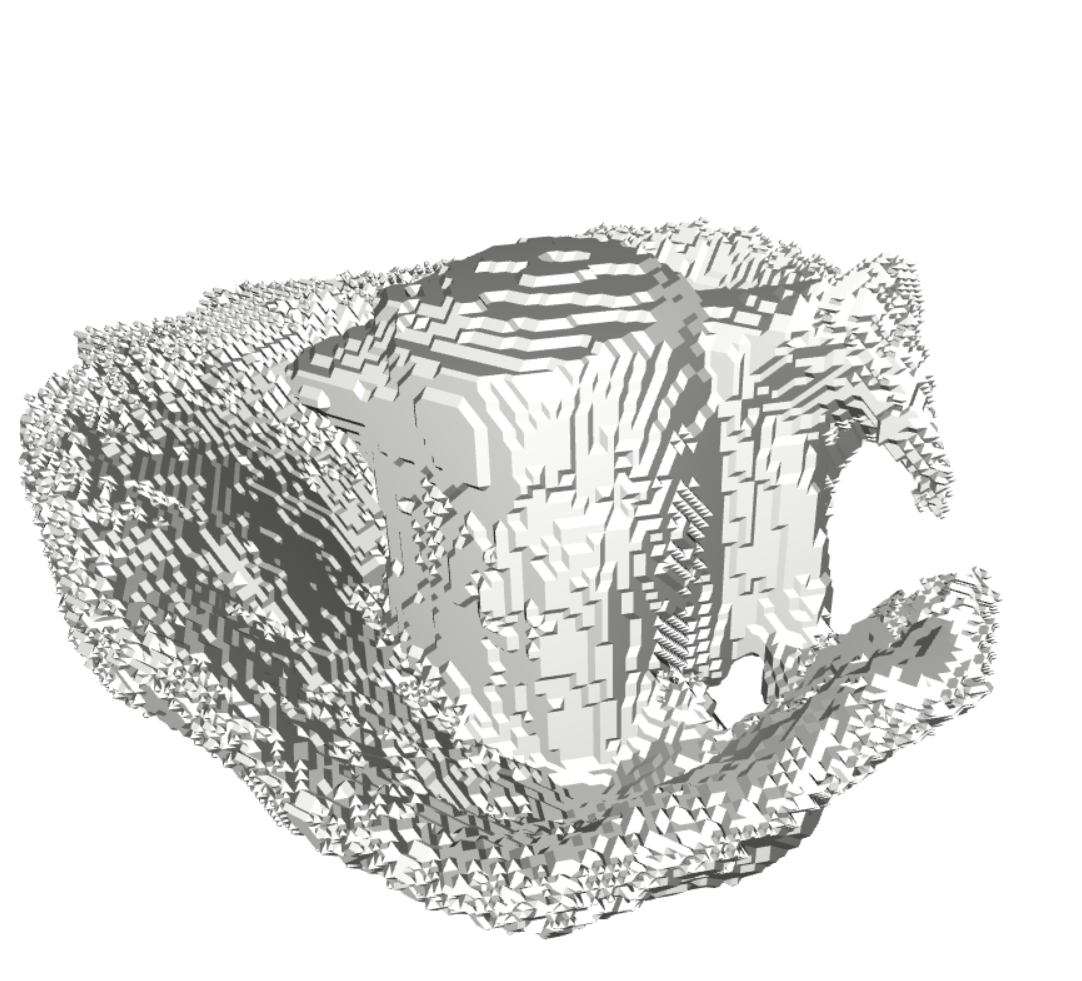}} &
        \makecell{\includegraphics[height=0.1\textwidth]{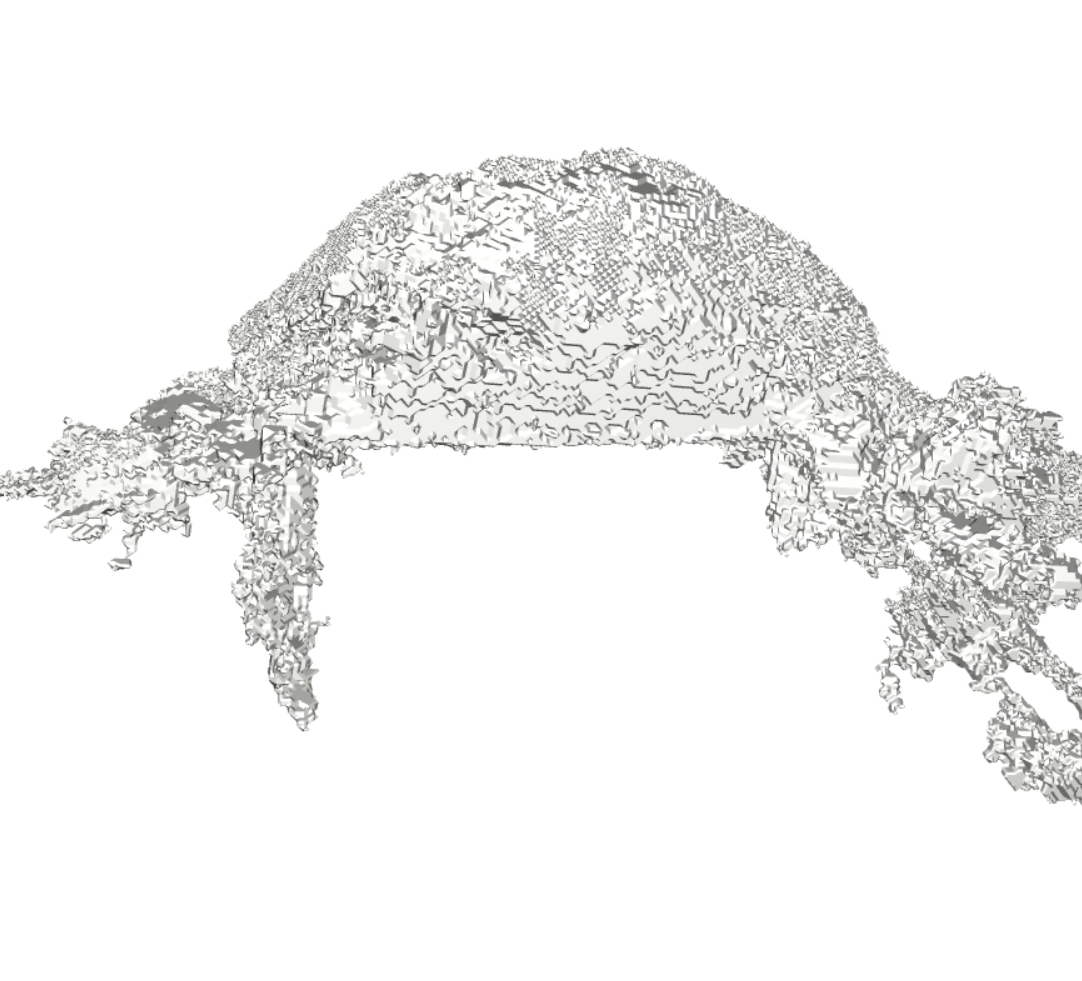}}&
        \makecell{\includegraphics[height=0.1\textwidth]{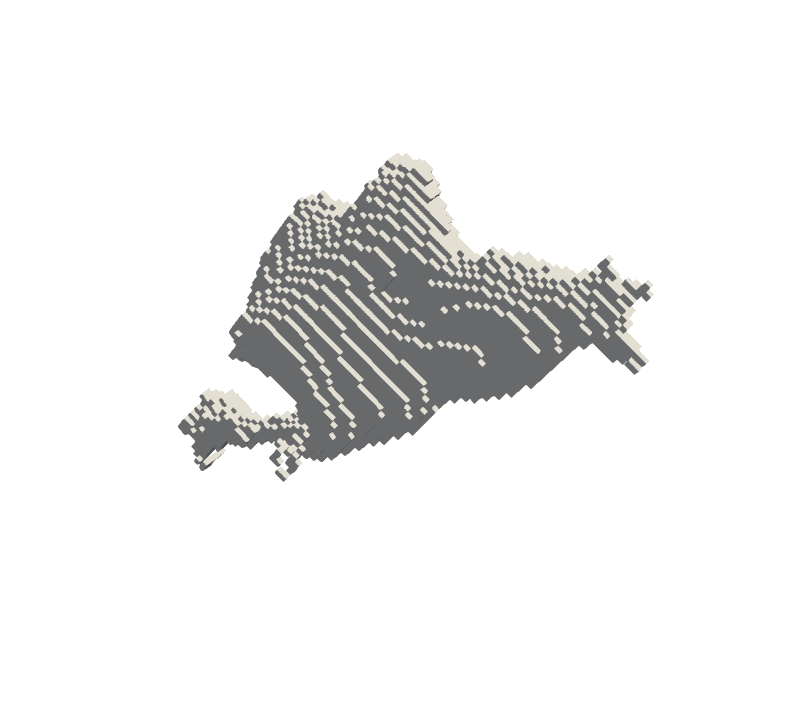}}&
        \makecell{\includegraphics[height=0.1\textwidth]{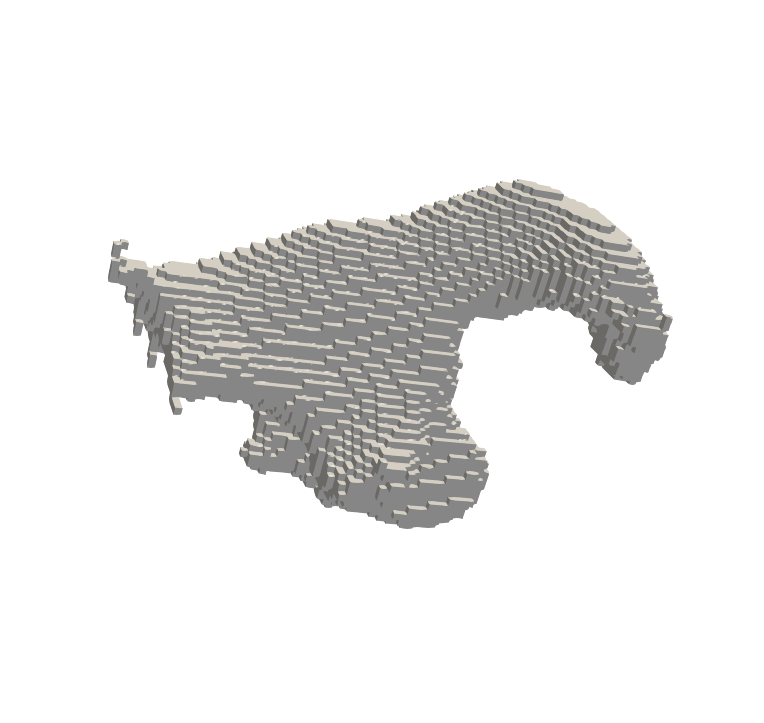}}
        \\[-5pt]
         \midrule
        \makecell{Thermoxels \\ (Ours)} &
        \makecell{\includegraphics[height=0.1\textwidth]{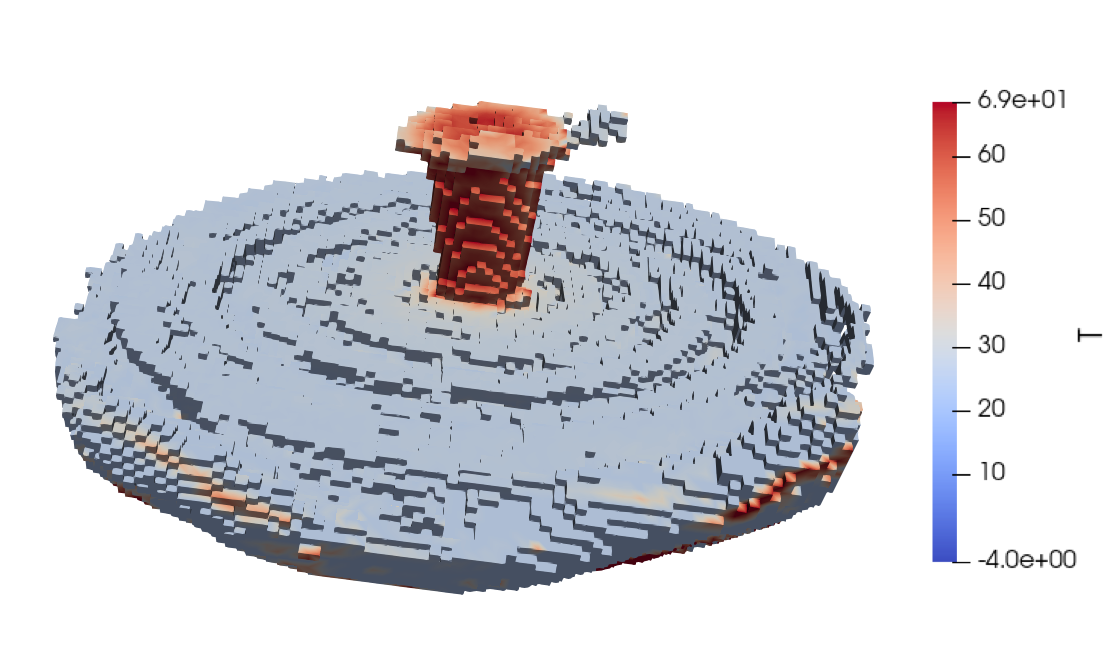}}      &
        \makecell{\includegraphics[height=0.1\textwidth]{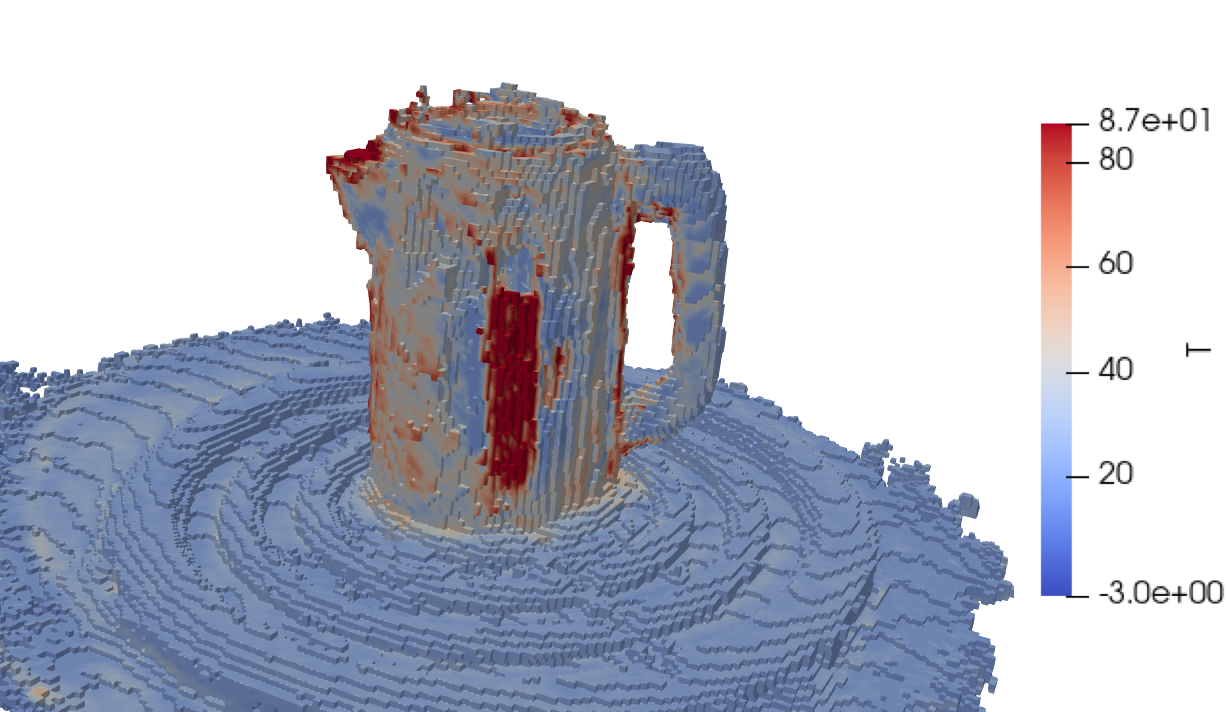}} &
        \makecell{\includegraphics[height=0.13\textwidth]{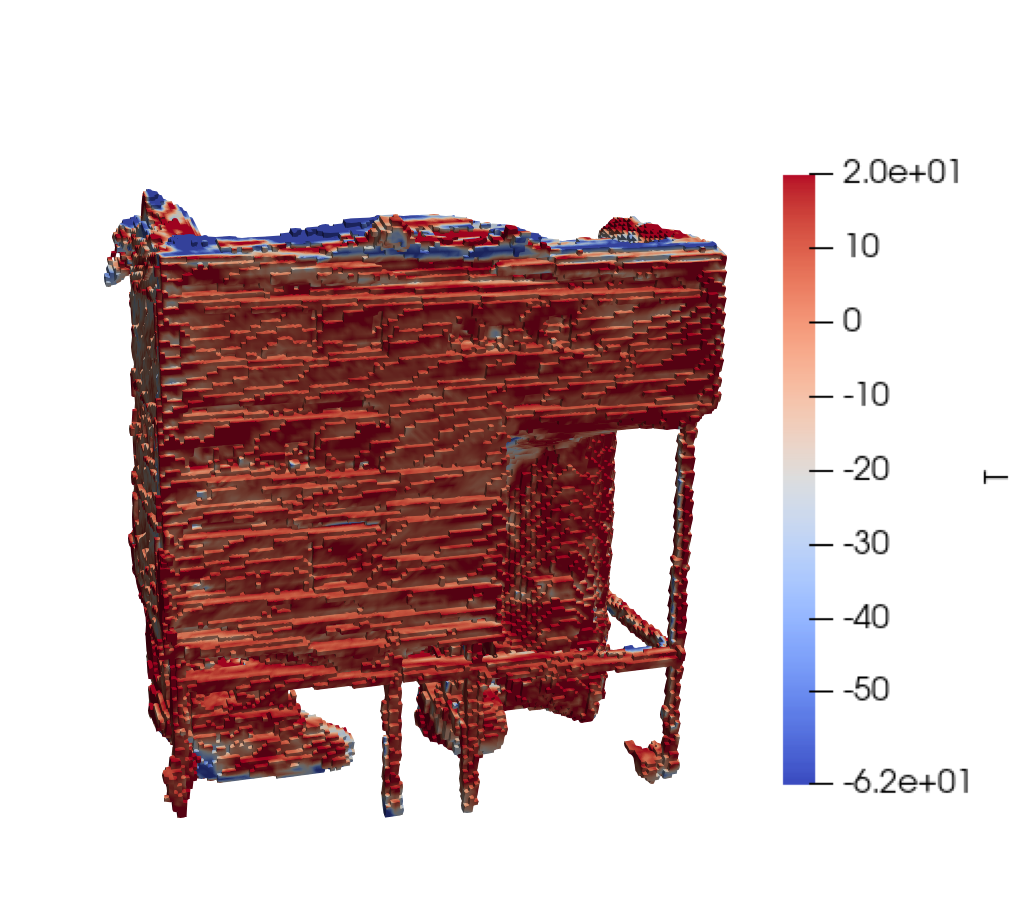}}&
        \makecell{\includegraphics[height=0.13\textwidth]{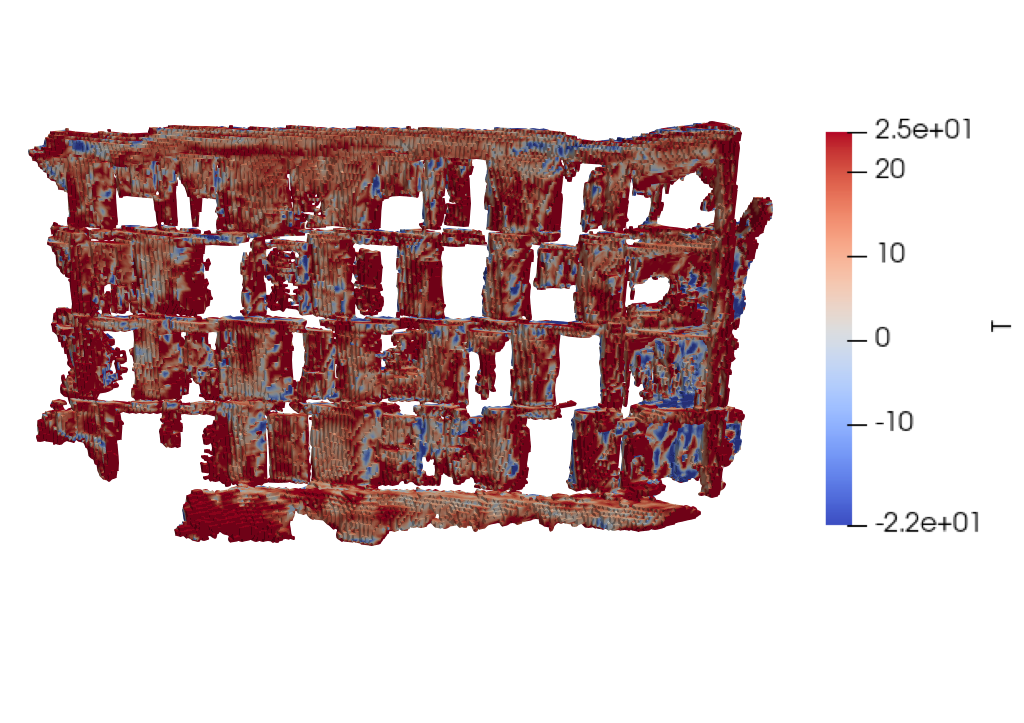}}&
        \makecell{\includegraphics[height=0.11\textwidth]{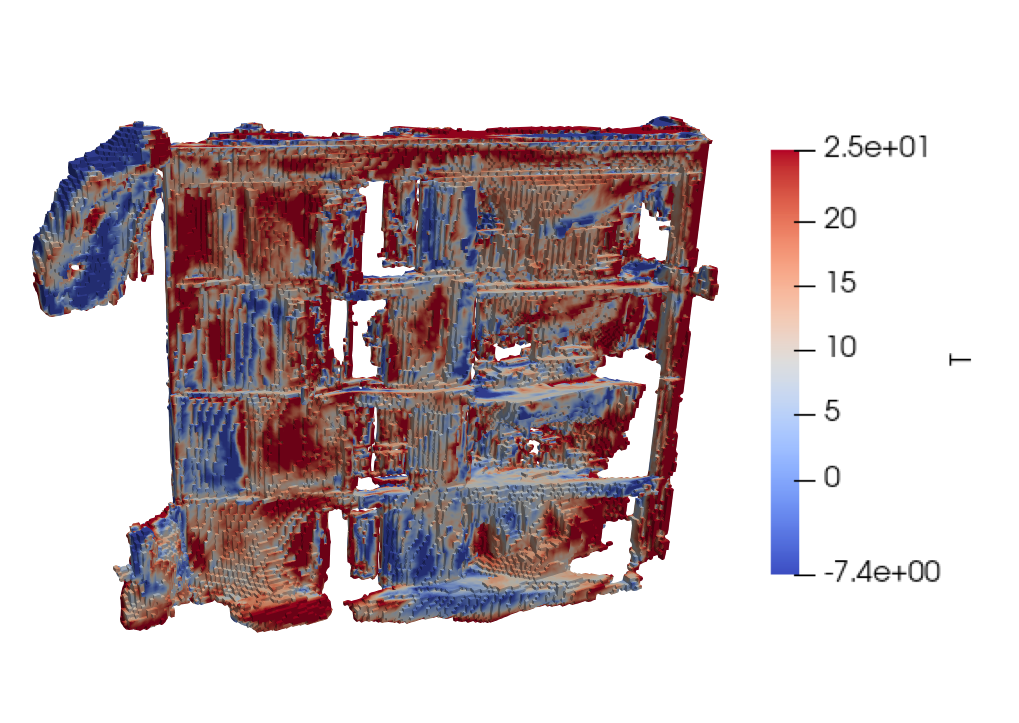}}                \\[-5pt]
        \makecell{Thermoxels \\ After FEA} &
        \makecell{\includegraphics[height=0.11\textwidth]{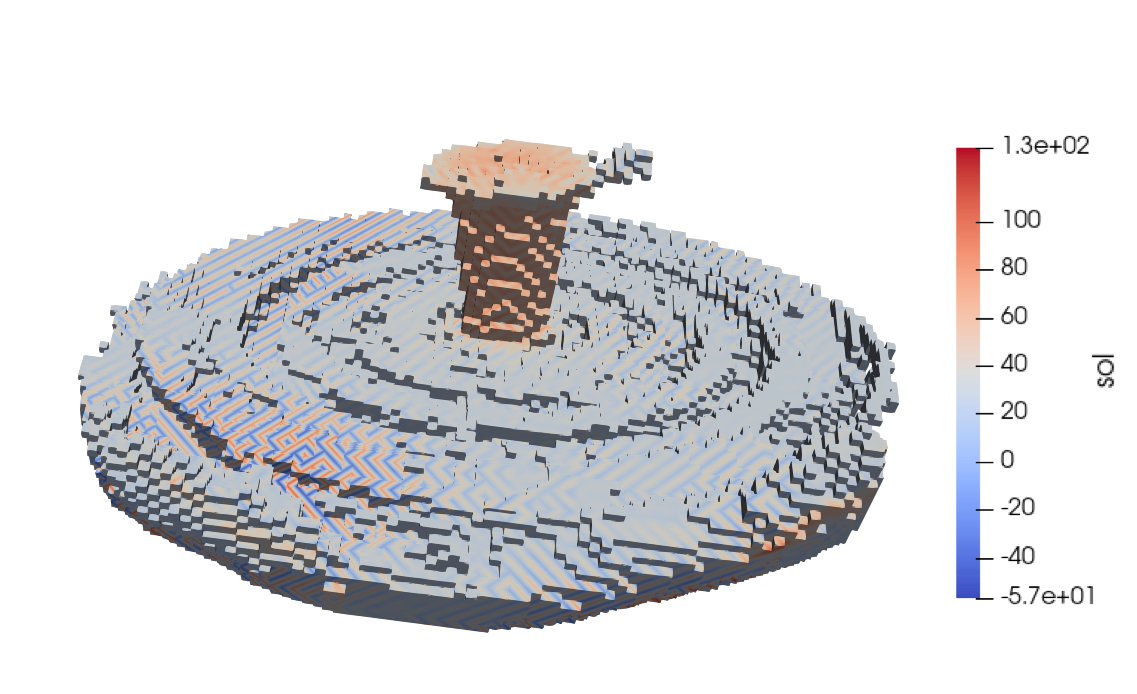}}      &
        \makecell{\includegraphics[height=0.11\textwidth]{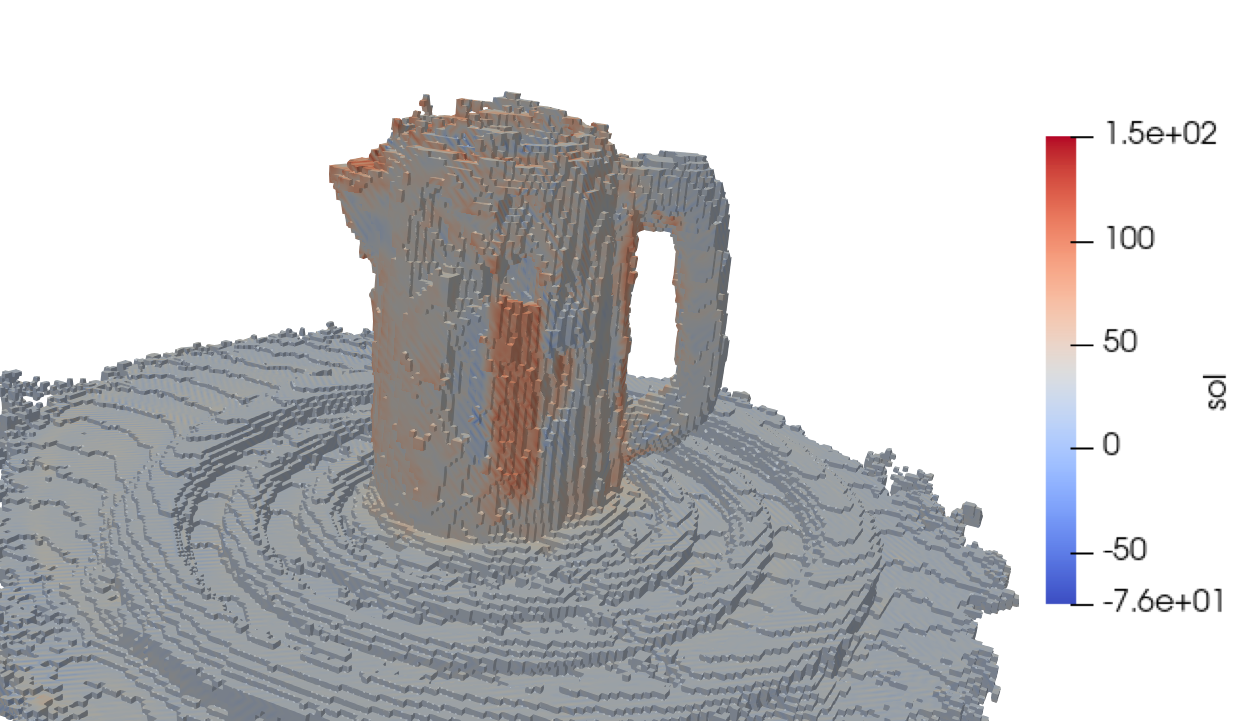}} &
        \makecell{\includegraphics[height=0.13\textwidth]{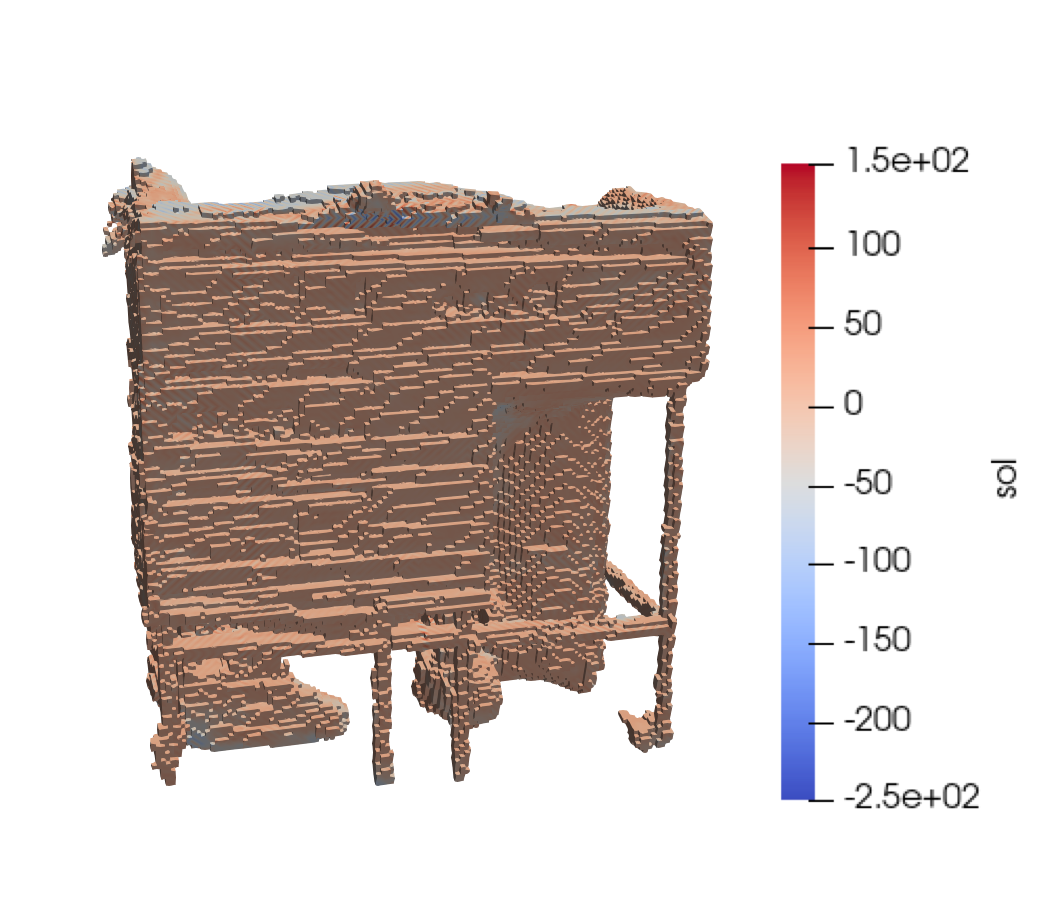}}&
        \makecell{\includegraphics[height=0.13\textwidth]{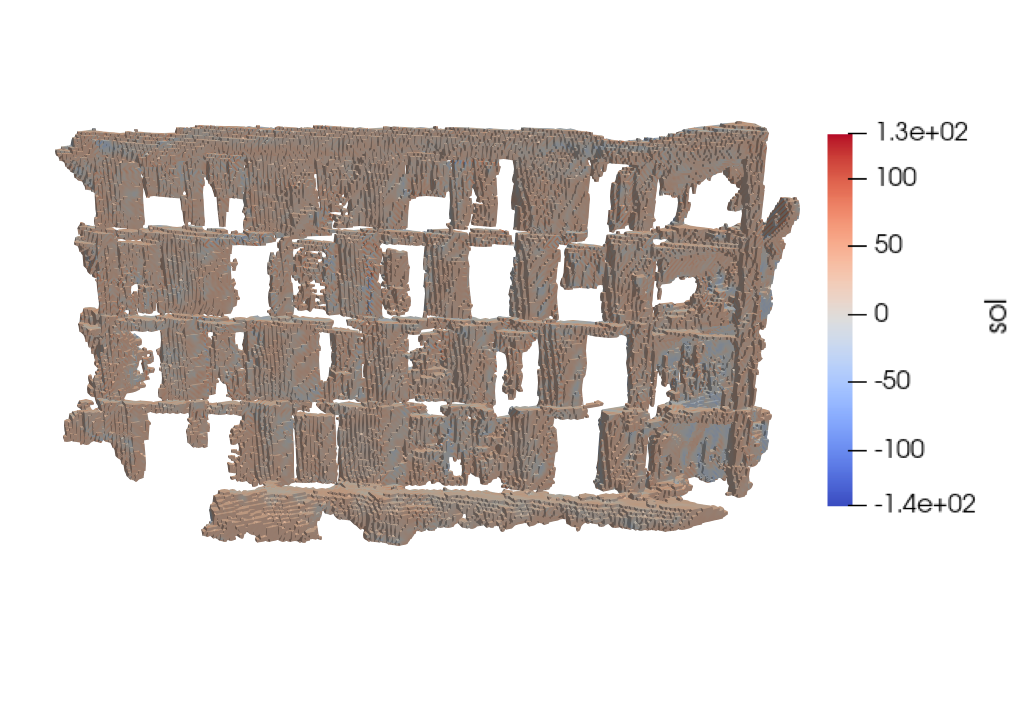}}&
        \makecell{\includegraphics[height=0.11\textwidth]{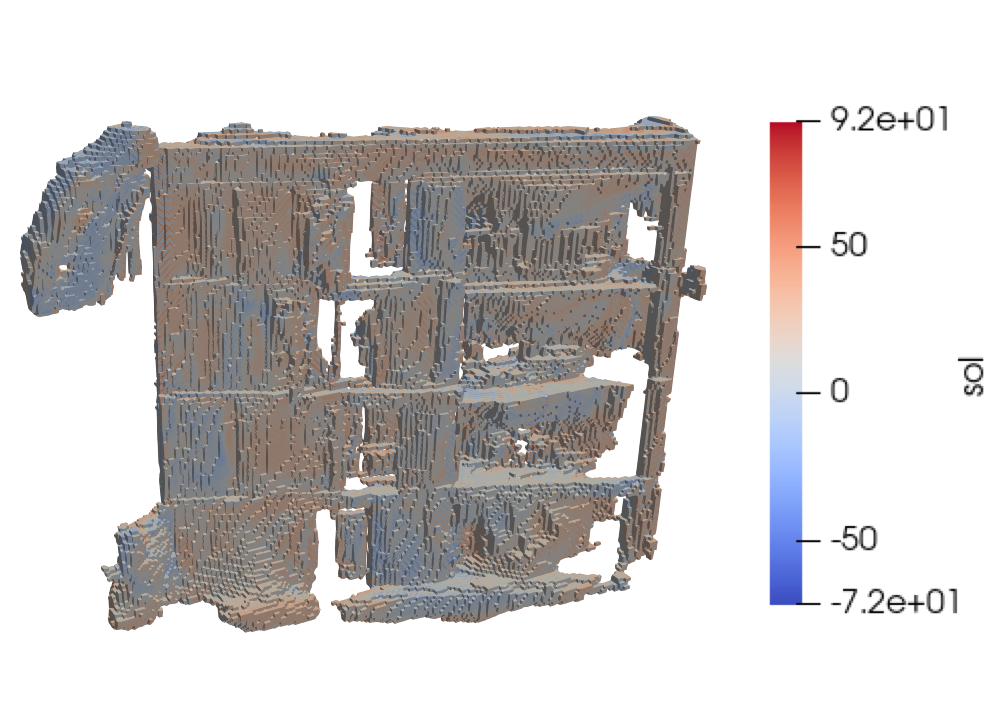} }               \\[-5pt]
        \bottomrule
    \end{tabular}}

    \caption{Examples of meshes reconstructed using Thermoxels and Plenoxels$_t$.
    One can see that Thermoxels consistently leads to more accurate reconstructions than Plenoxels$_t$.
    Due to the spherical harmonics used by Plenoxels$_t$, the temperature on the surface varies depending on the viewing direction, which is why the exported meshes do not have associated temperatures.
    Blue depicts the minimum temperature (Celsius) of the scene, grey average, and red maximum. }
    \label{fig:comparison-vertical-tight}
\end{figure}

We observe that on small-scale everyday scenes, while both Thermoxels and Plenoxels$_t$ are able to reconstruct a volumetric mesh of the scene, Thermoxels reconstructions capture more details---e.g. in \cref{fig:comparison-vertical-tight}, the kettle handle is present in Thermoxels but not in Plenoxels$_t$, and the cup cap is present in Thermoxels but absent in Plenoxels$_t$.
In larger scenes, Thermoxels can reconstruct the scene with high accuracy, while Plenoxels$_t$ fail to infer the geometry of the scene from only thermal images, as seen in \cref{fig:comparison-vertical-tight}.
On the other hand, thanks to the RGB and thermal modalities, Thermoxels' volumetric meshes are geometrically consistent---the facade and circular column of building A spring are correctly reconstructed, while the facade of the Dorm 1 and 2 scenes are flat.
We observed that Thermoxels performs better with scenes that have a wider angle of view---e.g. Building A spring that includes the front facade and part of the side walls, as opposed to more ``flat'' scenes such as Dorm 1 and 2---leading to more details being rendered and a lower proportion of holes in the reconstruction.
Furthermore, highly reflective surfaces appear to often be estimated as empty regions with no density, leading to holes in the reconstruction.


To validate the compatibility of the generated mesh with FEA, we use JaxFEM as an FEA solver to run a heat conduction simulation on each mesh.
The simulation begins with the temperature field estimated by Thermoxels as the initial condition, and simulates heat conduction over 10 simulation steps, assuming a uniform and constant thermal conductivity over the whole scene.
As shown in \cref{fig:comparison-vertical-tight}, the simulation successfully converges to a solution without requiring any manual post-processing of the mesh, demonstrating its compatibility with FEA simulation.
While out of the scope of this paper, in future work, we plan to use realistic thermal conductivity parameters and to build a realistic volumetric model by hand to obtain ground-truth simulation results to evaluate the Thermoxels-based simulation quantitatively.

\subsection{Thermal reconstruction results}
\label{sec:eval:synthesized}

\begin{table*}[t]
       \centering
       \caption{Evaluation of thermal reconstruction in terms of PSNR, SSIM and MAE.
              The metrics poorly reflect the 3D reconstruction quality; although ThermoNeRF and Plenoxels$_t$ scored higher, neither produces realistic 3D models (see \cref{sec:eval:fea}).
              Additionally, the higher temperature errors of Building A Spring, Dorm 1 et 2 are primarily caused by the lack of background optimization, which are filled with the scene's minimum temperature, driving up the error (e.g. due to the sky being visible on the image or holes in the meshes as in \cref{fig:comparison-vertical-tight}).
       }
       \resizebox{\textwidth}{!}{
              \begin{tabular}{lccccccccccccc}
                     \toprule
                     Metric & Method                 & \specialcell{Heated Water                                                                 \\ Cup} & \specialcell{Heated \\ Water Kettle} & \specialcell{Melting \\ Ice Cup} & \specialcell{Building \\ (Spring)} & \specialcell{Building \\ (Winter)} & \specialcell{Double \\ Robot} & \specialcell{Exhibition \\ Building} & \specialcell{Dorm 1} & \specialcell{Dorm 2} \\
                     \midrule
                     \multirow{3}{*}{PSNR$\uparrow$}
                            & ThermoNeRF             & 32.05                     & 34.04 & 32.24 & 26.63 & 28.75 & 30.75 & 33.79 & 34.10 & 29.94 \\
                            & Plenoxels$_{\text{t}}$ & 31.93                     & 34.40 & 42.05 & 35.24 & 34.62 & 33.29 & 27.50 & 39.24 & 35.5  \\
                            & Thermoxels             & 29.78                     & 30.37 & 19.99 & 19.84 & 22.56 & 24.16 & 18.78 & 10.79 & 13.85 \\

                     \midrule
                     \multirow{3}{*}{SSIM$\uparrow$}
                            & ThermoNeRF             & 0.92                      & 0.94  & 0.98  & 0.92  & 0.88  & 0.95  & 0.97  & 0.96  & 0.95  \\
                            & Plenoxels$_{\text{t}}$ & 0.92                      & 0.98  & 0.96  & 0.96  & 0.93  & 0.92  & 0.97  & 0.98  & 0.97  \\
                            & Thermoxels             & 0.83                      & 0.92  & 0.88  & 0.90  & 0.82  & 0.75  & 0.86  & 0.60  & 0.72  \\

                     \midrule
                     \multirow{3}{*}{MAE$\downarrow$}
                            & ThermoNeRF             & 2.10                      & 2.76  & 1.57  & 1.88  & 0.66  & 0.91  & 0.31  & 0.38  & 0.75  \\
                            & Plenoxels$_{\text{t}}$ & 0.87                      & 1.41  & 0.11  & 1.36  & 0.40  & 0.49  & 1.00  & 0.32  & 0.39  \\
                            & Thermoxels             & 0.99                      & 1.15  & 1.74  & 5.29 & 1.27  & 0.74  & 1.29  & 8.87  & 4.21  \\
                     \bottomrule
              \end{tabular}
       }
       \label{table:thermal-results}
\end{table*}

Thermal reconstruction results for  scenes in the ThermoScenes dataset are shown in \cref{table:thermal-results}.
We evaluate Thermoxels against Plenoxels$_t$ and ThermoNeRF~\cite{hassan_thermonerf_2024}, a NeRF-based RGB+thermal novel view synthesis method, unable to generate volumetric meshes.
For RGB images, the peak signal-to-noise ratio (PSNR), derived from the mean squared error, and structural similarity index measure (SSIM), derived from correlations between the images, are used as metrics.
For thermal image synthesis, we evaluate the mean absolute error (MAE) over the entire image.
It should be noted that, since we aim to obtain volumetric meshes compatible with FEA, mesh geometry and temperature reconstruction are more critical than novel view synthesis.

Despite exhibiting inferior results on image reconstruction metrics (PSNR and SSIM), our model achieves competitive performance on thermal MAE.
Notably, Plenoxels$_t$ obtains the best scores on most metrics even though the method is unable to reconstruct accurately the geometry of the scene, as shown in \cref{sec:eval:fea}.
We hypothesize that it is due to the encoding of the temperatures using view-dependent spherical harmonics and the lack of contrast in thermal images.
This combination enables Plenoxels$_t$ to synthesize accurate images from incorrect geometries, adapting the temperature values based on the camera pose, instead of the sampling position $p_i$ in space.
On the other hand, Thermoxels uses the high-contrast RGB images to estimate the geometry onto which the temperature is optimized, constraining the temperature field and thus leading to more accurate geometry but worse image synthesis.
Compared to ThermoNeRF, which fully decouples RGB and thermal estimation, Thermoxels' temperature estimation can be impacted by the RGB modality as seen in \cref{fig:comparison-vertical-tight}, where, while the kettle temperature is correctly estimated, the temperature of the white band of the dart board is influenced by the RGB modality.

\section{Conclusion}

In this work, we introduced Thermoxels, a volumetric thermal novel view synthesis method capable of generating a structured 3D thermal field representation.
Thermoxels enables the extraction of a volumetric mesh suitable for downstream simulation tasks, and thermal simulations were conducted using Themoxels' volumetric meshes and temperature estimates as initial conditions.
However, more work is needed to improve the robustness of the reconstruction of reflective surfaces, flat geometries and to avoid information leaking between modalities.
Future work will also look into adding realistic material properties to the simulation.

\section*{References}

\bibliography{references}

\providecommand{\newblock}{}
\begin{thebibliography}{10}
\expandafter\ifx\csname url\endcsname\relax
  \def\url#1{{\tt #1}}\fi
\expandafter\ifx\csname urlprefix\endcsname\relax\def\urlprefix{URL }\fi
\providecommand{\eprint}[2][]{\url{#2}}

\bibitem{european2020renovation}
Commission E 2020 A renovation wave for europe—greening our buildings, creating jobs, improving lives {\em Official Journal of the European Union\/}  26

\bibitem{sanhudo_building_2018}
Sanhudo L, Ramos N~M, Poças~Martins J, Almeida R~M, Barreira E, Simões M~L and Cardoso V Building information modeling for energy retrofitting – a review {\bf 89} 249--260 ISSN 13640321 \urlprefix\url{https://linkinghub.elsevier.com/retrieve/pii/S1364032118301503}

\bibitem{mildenhall2020nerf}
Mildenhall B, Srinivasan P~P, Tancik M, Barron J~T, Ramamoorthi R and Ng R 2020 Nerf: Representing scenes as neural radiance fields for view synthesis {\em ECCV\/}

\bibitem{chen_survey_2024}
Chen G and Wang W A survey on 3d gaussian splatting \urlprefix\url{http://arxiv.org/abs/2401.03890}

\bibitem{wang_neus_2021}
Wang P, Liu L, Liu Y, Theobalt C, Komura T and Wang W 2021 {NeuS}: Learning neural implicit surfaces by volume rendering for multi-view reconstruction \urlprefix\url{https://arxiv.org/abs/2106.10689}

\bibitem{xu_exploiting_2025}
Xu C, Mielle M, Laborde A, Waseem A, Forest F and Fink O 2025 Exploiting semantic scene reconstruction for estimating building envelope characteristics {\em Building and Environment\/} {\bf 275} 112731 ISSN 03601323 \urlprefix\url{https://linkinghub.elsevier.com/retrieve/pii/S0360132325002136}

\bibitem{xie_physgaussian_2023}
Xie T, Zong Z, Qiu Y, Li X, Feng Y, Yang Y and Jiang C {PhysGaussian}: Physics-integrated 3d gaussians for generative dynamics \urlprefix\url{http://arxiv.org/abs/2311.12198}

\bibitem{carrasco_building_2023}
Carrasco C~A, Lombillo I, Balbás F~J, Aranda J~R and Villalta K Building information modeling ({BIM} 6d) and its application to thermal loads calculation in retrofitting {\bf 13} 1901 ISSN 2075-5309 \urlprefix\url{https://www.mdpi.com/2075-5309/13/8/1901}

\bibitem{cho_3d_2015}
Cho Y~K, Ham Y and Golpavar-Fard M 3d as-is building energy modeling and diagnostics: A review of the state-of-the-art {\bf 29} 184--195 ISSN 14740346 \urlprefix\url{https://linkinghub.elsevier.com/retrieve/pii/S1474034615000312}

\bibitem{mildenhall2021nerf}
Mildenhall B, Srinivasan P~P, Tancik M, Barron J~T, Ramamoorthi R and Ng R 2021 Nerf: Representing scenes as neural radiance fields for view synthesis {\em Communications of the ACM\/} {\bf 65} 99--106

\bibitem{kerbl20233d}
Kerbl B, Kopanas G, Leimk{\"u}hler T and Drettakis G 2023 3d gaussian splatting for real-time radiance field rendering. {\em ACM Trans. Graph.\/} {\bf 42} 139--1

\bibitem{fridovich2022plenoxels}
Fridovich-Keil S, Yu A, Tancik M, Chen Q, Recht B and Kanazawa A 2022 Plenoxels: Radiance fields without neural networks {\em Proceedings of the IEEE/CVF conference on computer vision and pattern recognition\/} pp 5501--5510

\bibitem{li2022vox}
Li H, Yang X, Zhai H, Liu Y, Bao H and Zhang G 2022 Vox-surf: Voxel-based implicit surface representation {\em IEEE Transactions on Visualization and Computer Graphics\/} {\bf 30} 1743--1755

\bibitem{hassan_thermonerf_2024}
Hassan M, Forest F, Fink O and Mielle M {ThermoNeRF}: Multimodal neural radiance fields for thermal novel view synthesis \urlprefix\url{http://arxiv.org/abs/2403.12154}

\bibitem{chen2024thermal3d}
Chen Q, Shu S and Bai X 2024 Thermal3d-gs: Physics-induced 3d gaussians for thermal infrared novel-view synthesis {\em European Conference on Computer Vision\/} (Springer) pp 253--269

\bibitem{zeise_temperature_2016}
Zeise B and Wagner B Temperature correction and reflection removal in thermal images using 3d temperature mapping: {\em Proceedings of the 13th International Conference on Informatics in Control, Automation and Robotics\/} pp 158--165 ISBN 978-989-758-198-4 \urlprefix\url{http://www.scitepress.org/DigitalLibrary/Link.aspx?doi=10.5220/0005955801580165}

\bibitem{chai_bladder_voxel_fem_2011}
Chai X, van Herk M, Hulshof M~C~C~M and Bel A 2012 A voxel-based finite element model for the prediction of bladder deformation {\em Medical Physics\/} {\bf 39} 55--65 \urlprefix\url{https://aapm.onlinelibrary.wiley.com/doi/abs/10.1118/1.3668060}

\end{thebibliography}

\end{document}